\ifdefined\XeTeXversion\else\ifdefined\pdfoutput
\pdfoutput=1
\fi\fi
\documentclass[11pt]{article}
\usepackage[T1]{fontenc}
\usepackage[utf8]{inputenc}
\usepackage{textcomp}
\usepackage{newtxtext,newtxmath}
\usepackage[letterpaper,margin=1in]{geometry}
\usepackage{graphicx}
\usepackage{float}
\usepackage{booktabs}
\usepackage{array}
\usepackage{tabularx}
\usepackage{enumitem}
\usepackage{xcolor}
\usepackage{listings}
\usepackage{tikz}
\usetikzlibrary{positioning,arrows.meta}
\usepackage[font=small,labelfont=bf]{caption}
\usepackage[hidelinks]{hyperref}

\hypersetup{
  pdftitle={Can an LLM Actually Reason and Program about Neural Network's Weights or is it just memorization on steroids? Exploring Weight-Space Programming},
  pdfauthor={Nazmus Sakib Ahmed}
}

\newcolumntype{Y}{>{\raggedright\arraybackslash}X}

\title{Can an LLM Actually Reason and Program about Neural Network's Weights or is it just memorization on steroids?\\[0.3em] Exploring ``Weight-Space Programming''}
\author{Nazmus Sakib Ahmed\\ BRAC University, Dhaka, Bangladesh\\ \texttt{nazmus.sakib@bracu.ac.bd}}
\date{November 2026}

\begin{document}
\maketitle

\begin{abstract}
Large language models (LLMs) have recently been reported to read and write the raw weights of neural networks directly, a capability termed \emph{weight-space programming}. We ask whether the reading side of this capability reflects reasoning about weight flows or a combination of local arithmetic at the network's edges and memorized priors over public datasets. We test Claude Opus 5.5 on small convolutional networks of about 125,000 parameters trained on five image datasets, providing every weight as plain text with no tools, no forward passes, and no dataset or class names. The model performs accurate local calculations: batch-normalization statistics lead it to infer about three optimizer steps on networks stopped after three updates, and output-layer sign patterns recover the vehicle/animal split of CIFAR-10. The sessions report reading the first and last layers closely and sampling the middle, and several claims built on partial tracing fail verification, although we did not run a direct forward-pass test. Semantic judgments lean on familiar datasets: the model names CIFAR-10 for CIFAR-10 networks but also for all three fast-food networks, relabels KMNIST as MNIST, reads a 99.57\%-accurate SAT-6 network as CIFAR-10 animal photos, and calls ``frog'' the output favored by a feature that fires most on trees. Permuting CIFAR-10's labels preserves the output split at the new indices, but every class pair the model interprets is then wrong, indicating reliance on remembered label order. These are exploratory results from mostly single runs. We conclude that zero-shot weight reading by this model is better described as local arithmetic plus dataset priors than as weight-flow tracing, and we provide a protocol for auditing future weight-space claims.
\end{abstract}


\section{Introduction}

Can a language model be handed the raw parameters of a neural network---a long list of floating-point numbers---and say what the network was trained to do? A recent technical report by Takehi et al.~\cite{takehi} argues yes. They report that GPT-6 Astra can read and write weights directly, without gradient descent: it names the game a small policy network was trained on, fixes a failing control network by editing 3.5\% of its weights, writes game-playing networks from scratch, and improves Qwen3.5-0.8B on MATH-500 from 7.9\% to 48.1\% by editing 0.016\% of its weights. They call this \emph{weight-space programming} (WSP).

These results are striking because they suggest that a general-purpose model can operate on the parameter space of another network as if it were text. If the reading side of this claim holds, the implications for interpretability and model auditing are substantial: one could inspect a model's weights for hidden capabilities without designing inputs or running forward passes. The claim also has a deflationary alternative. A model that has seen many public datasets and frameworks may recognize the \emph{footprint} of a familiar training setup---input size, output count, initialization scale---and fill in the rest from priors. A network with $32\times32$ inputs and ten outputs already looks like CIFAR-10 before a single weight is read.

\begin{figure}[t]
  \centering
  \includegraphics[width=0.85\linewidth]{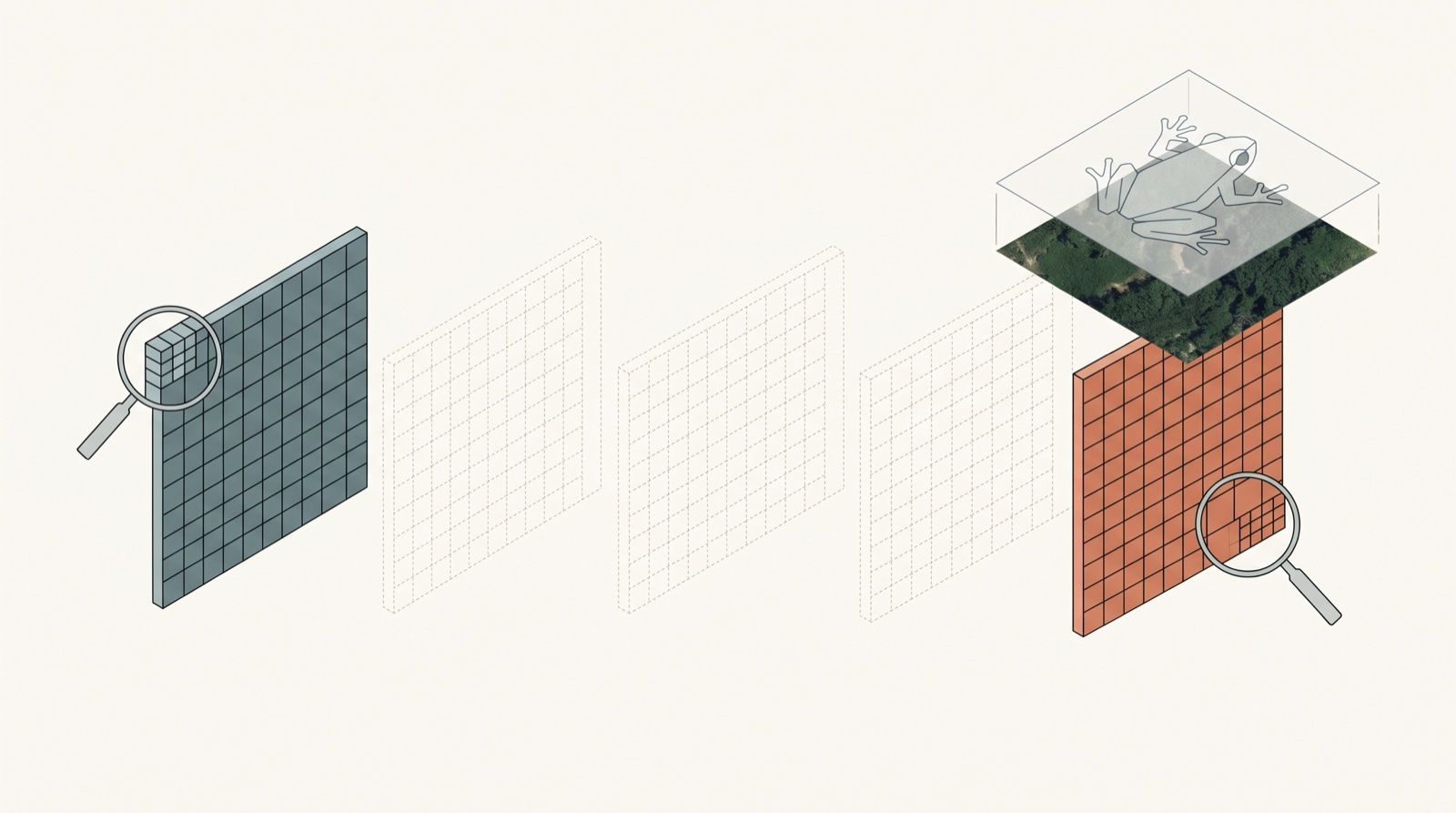}
  \caption{Conceptual illustration of the central finding. The model inspects the outer layers of a trained network and fills the semantic gap with a familiar template; here, a geometric frog is imposed on a satellite forest patch. The illustration is not a literal rendering of the experiments.}
  \label{fig:concept}
\end{figure}

We test this alternative on the reading side of WSP. We give Claude Opus 5.5 the complete weights of small convolutional networks, as plain text, in a fresh session with no tools, no memory, and no ability to run the network. We train the networks ourselves on five image datasets: CIFAR-10, KMNIST, Cats vs Dogs, a fast-food dataset, and SAT-6. We vary the architecture, the dataset, the label order, and the network's test accuracy. We first ask which dataset a network was trained on. Then, with a revised prompt on a subset of conditions, we ask the model to describe the images and the meaning of each output. We check specific claims against the actual weights and, where relevant, against the real data.

Three findings emerge. First, the model is accurate at local arithmetic. It infers that a network stopped after three updates took about three Adam steps, using batch-norm running statistics and PyTorch defaults, and it finds the output-layer sign pattern that separates CIFAR-10 vehicles from animals. Second, we find no sign that it traces the intermediate computations connecting inputs to outputs. The sessions report reading the edges closely and sampling the middle, and several claims made from partial tracing fail our checks. Third, its semantic judgments lean on familiar datasets. It gives CIFAR-10 for CIFAR-10 networks but also for fast-food networks, collapses KMNIST to MNIST, reads a near-perfect SAT-6 network as CIFAR-10 animal photos, and calls ``frog'' the output favored by a feature that fires most on trees. When we permute \mbox{CIFAR-10}'s labels, the model still finds the output split at the new indices, but every class pair it interprets is wrong. The arithmetic survives; the semantics do not.

These findings do not mean that LLMs have no mechanistic interpretability capability. The model does recover genuine structure from the weights, and that recovery is itself a form of interpretation. The pattern instead suggests that the capability is bounded by large-scale memory and pattern recognition over familiar architectures, datasets, and framework conventions: the model recognizes weight signatures it has seen before, and we found no evidence that it generalizes that recognition into tracing an arbitrary network's computation.

\paragraph{Contributions.} We make three contributions.
\begin{enumerate}[leftmargin=*]
  \item \textbf{A controlled audit of zero-shot weight reading.} We test whether a general-purpose LLM, without tools or fine-tuning, can infer dataset and class semantics from the raw weights of small CNNs trained on five datasets.
  \item \textbf{Controls aimed at separating local arithmetic from dataset priors.} We include a label permutation, a shuffled-weight control, an architecture comparison, datasets chosen to differ in familiarity, and an accuracy comparison. The label permutation is the key diagnostic: it preserves the learned output-layer structure while removing the mapping from indices to class names.
  \item \textbf{A characterization of the failure mode and a reusable protocol.} We show that the model's checked successes are concentrated at the network's edges, that its partial-tracing claims are unreliable, and that its semantic errors follow familiar public datasets. We provide the prompts, scoring rules, and per-run results so the protocol can be reused.
\end{enumerate}

\paragraph{Scope.} We test one model (Claude Opus 5.5), not GPT-6 Astra, and one task family (small image classifiers). We do not test weight editing. This is an exploration of the reading mechanism, not a replication of Takehi et al.~\cite{takehi}, and it does not refute their report: our results do not establish that their reading, writing, or editing results fail. A preliminary version of this work appeared as a blog post.\footnote{\url{https://sakibbai.com/posts/2026-11-05-llm-weight-space-programming/}}

\section{Related Work}

\subsection{Interpretability from activations and circuits}

Mechanistic interpretability reverse-engineers the computations of trained networks from their internal states. Circuit analyses of vision models identify features and the circuits that combine them~\cite{olah}; mathematical frameworks for transformer circuits decompose attention heads and residual-stream computations~\cite{elhage}, including the indirect object identification circuit in GPT-2 small~\cite{wangR}. Sparse dictionary learning decomposes activations into sparse, more interpretable features~\cite{bricken,cunningham}, and attribution-graph methods extend circuit tracing to larger models~\cite{ameisen}. Reviews catalogue both the methods and the open problems~\cite{bereska,sharkey}.

These methods require activations from forward passes, access to internal states, or both. The setting we study removes both: the model sees only a text serialization of the weights, and it cannot run the network. This makes the task closer to reading a program than to observing a computation.

\subsection{Reading weights directly}

A separate line analyzes weights without data. WeightWatcher uses spectral properties of weight matrices to predict generalization trends without access to training or test data~\cite{martin}. Equivariant architectures learn functions on weight spaces~\cite{navon}, hyper-representations learn representations of weights~\cite{schurholt}, and generative models produce checkpoints or weights~\cite{peebles,wangJ,wangK}. AdderBoard is a community leaderboard for the smallest transformer that adds two 10-digit numbers, with trained and hand-coded entries, some built with coding agents~\cite{papailiopoulos}. These approaches either train a model on weight space, constrain the architecture, or rely on search; to our knowledge, none of them tests whether a general-purpose LLM can read an arbitrary small network zero-shot.

\subsection{LLMs as interpreters}

Language models have been used as explainers of model internals. Bills et al.~\cite{bills} use GPT-4 to produce natural-language explanations of individual neurons. Activation Oracles train LLMs to answer arbitrary questions about activations~\cite{karvonen}, and natural-language autoencoders learn unsupervised explanations of activations~\cite{fraser}. Closest to our setting, Weight Oracles fine-tune language models to read raw weights, using a staged curriculum and an external chain of computation that delegates parameter-free operations to code; they report up to 99\% holdout accuracy when simulating small transformers~\cite{kabra}. Diff Interpretation Tuning trains adapters that describe finetuning-induced weight differences~\cite{goel}.

These systems are trained for interpretation and, in Weight Oracles, given external computation. Our study asks what an untrained, general-purpose model does with weights alone, with no tools. The distinction matters for the interpretation of WSP: a capability demonstrated by a fine-tuned specialist is not evidence that a general model reasons about weights natively.

\subsection{Editing and optimization in weight space}

ROME and MEMIT locate and edit factual associations in transformer weights~\cite{meng2022,meng2023}, and refusal behavior can be mediated by a single direction~\cite{arditi}. HyperNetworks generate weights with an auxiliary network~\cite{ha}, and forward-only methods such as MeZO optimize without backpropagation~\cite{malladi}. These results show that weights are actionable; they do not show that an LLM can infer semantics from weights without training. Our audit isolates the reading step.

\subsection{What models reveal about their training data}

Membership inference determines whether an example was used in training~\cite{shokri}; property inference recovers aggregate properties of the training set~\cite{ganju}; dataset inference attributes a model to a dataset~\cite{maini}. LLMs memorize training text~\cite{carliniExtract,carliniQuant} and are partly calibrated about what they know~\cite{kadavath}.

These mechanisms concern information carried by the \emph{target} model. Our hypothesis is complementary: the \emph{interpreter} LLM's prior over public datasets may supply semantic labels that the weights themselves do not establish. The distinction is important because it means that a correct dataset guess is not, by itself, evidence that the weights were understood.

\subsection{Positioning}

We audit the reading claim of WSP~\cite{takehi} with a different model and a controlled task family, and we add controls that dataset familiarity makes necessary: label permutation, an architecture comparison, datasets chosen to differ in familiarity, and an accuracy comparison. Our result is exploratory and methodological: zero-shot reading by this model is better described as local arithmetic plus priors than as weight-flow tracing.

\section{Background: Weight-Space Programming}

Takehi et al.~\cite{takehi} define weight-space programming as a general-purpose LLM reading and writing a network's weights directly, around a natural-language specification, without an optimizer or gradient descent. Their report spans three capabilities. \emph{Reading:} a model names the game a small policy network was trained to play, the shape a diffusion model generates, and the language a tiny transformer was trained on. It could not tell which of two language models was sycophantic from the weights alone, but it could when allowed to inspect their attention patterns. \emph{Writing:} a model writes, from scratch, small game-playing networks that outperform PPO-trained networks, and a Go-playing network at the 1.5-million-parameter scale. \emph{Editing:} a model fixes a failing control network by editing 3.5\% of its weights, and improves Qwen3.5-0.8B on \mbox{MATH-500} from 7.9\% to 48.1\% by editing 131,072 of its 0.8 billion weights (0.016\%), with instruction-following performance also improving.

Two features of that setup matter for interpreting the results. First, in the game experiment the prompt labels what each input channel represents (agent, moving objects, reward targets, obstacles, previous frame), and the first-layer weights map directly onto grid coordinates. These labels give the model a semantic anchor. Our prompts provide layer shapes and raw numbers, name the input channels only as red, green, and blue, and give no class labels. Second, several of their experiments provide a Python environment or a budget of forward passes, although the game experiment does not. Our setting provides neither: the model cannot execute code, run the network, or query an environment.

We focus on the reading side because it bears on the stronger claims. If a model cannot reliably recover what a network was trained to do from weights alone, then reports of weight editing must be evaluated with care: an edit that improves a benchmark may have been found by search rather than by understanding. Conversely, an editing result may be real even if the reading account is prior-driven; our results do not settle that question.

\section{Experimental Setup}

\subsection{Networks and training}

We train three convolutional architectures. Each has three blocks of $3\times3$ convolution (padding 1), ReLU, $2\times2$ max pooling, and batch normalization~\cite{ioffe}, followed by a dense layer with ReLU and dropout ($p=0.3$), then an output layer. \textbf{Model 1} uses standard filter counts $(16,32,64)$ and a dense layer of 100 units. \textbf{Model 2} uses unusual counts $(13,33,77)$ and 78 units. \textbf{Random} uses $(62,68,25)$ and 175 units, drawn with a seeded random search from the 833 size combinations that match Model 1's 127,318 parameters and have no size divisible by 8 or 10. Because the output layer follows each dataset's class count, the networks have between 123,782 and 127,318 trained parameters.

We train with PyTorch~\cite{paszke} and Adam~\cite{kingma} at learning rate $10^{-3}$, batch size 128, learning-rate halving when validation loss plateaus (patience 3), early stopping with patience 10, up to 100 epochs, and seed 0. We keep the checkpoint with the lowest validation loss and use the test set only for final evaluation.

CIFAR-10 uses its standard normalization constants. KMNIST images are padded with zeros from $28\times28$ to $32\times32$, copied to three channels, and normalized with one mean and standard deviation computed on the padded training images. All other images are resized directly to $32\times32$ with bilinear interpolation, without cropping or preserving aspect ratio, and normalized per channel with training-split statistics. Training augmentation is a random crop with padding 4 and a horizontal flip; KMNIST uses the crop only, because a mirrored character is not a valid example of its class.

\subsection{Datasets}

\textbf{CIFAR-10} contains ten classes of $32\times32$ color images~\cite{krizhevsky}. Its standard label order places vehicles at outputs $\{0,1,8,9\}$ and animals at outputs $\{2,3,4,5,6,7\}$. \textbf{KMNIST} contains ten classes of Japanese characters in grayscale~\cite{clanuwat}; we use \nolinkurl{tanganke/kmnist} from Hugging Face. \textbf{Cats vs Dogs} is a public binary dataset; we use \nolinkurl{microsoft/cats_vs_dogs}, with 23,410 images. \textbf{Fast food} is a public ten-class food-image dataset; we use \nolinkurl{andrewsunanda/fast_food_image_classification}, with 3,000 images. \textbf{SAT-6} contains $28\times28$ satellite patches in six land-cover classes: barren land, building, grassland, road, trees, and water~\cite{basu}. We use \nolinkurl{jonathan-roberts1/SAT-6}, whose 81,000 RGB images are the original release's test split. CIFAKE~\cite{bird} appears in the results only as a model guess, not as a training dataset.

Table~\ref{tab:datasets} summarizes the configurations. CIFAR-10 and KMNIST use their standard test sets, with a stratified validation split from the training images. The other three datasets have one split, so we stratify 10\% of each class for test, 10\% for validation, and the rest for training. With 300 test images, fast-food accuracies carry a sampling margin of about $\pm5.6$ points.

SAT-6 is easy to learn: Model 2 reaches 89.14\% test accuracy from only 24 training images. To obtain a low-accuracy network, we use the same data, architecture, and settings, check validation accuracy after every update, and keep the state nearest a validation target of 55\%, a target set in advance. Accuracy jumped from 35.6\% to 76.6\% between the third and fourth updates, so the rule keeps the three-update network, whose test accuracy of 36.01\% is below the majority-class baseline.

\begin{table}[t]
  \centering
  \caption{Datasets, splits, and test accuracies. M1, M2, and R denote Model 1, Model 2, and Random. The SAT-6 low-accuracy network is the three-update model.}
  \label{tab:datasets}
  \small
  \setlength{\tabcolsep}{4pt}
  \begin{tabular}{@{}lrlrll@{}}
    \toprule
    Dataset & Classes & Train / val / test & Majority (\%) & Networks & Test accuracy (\%) \\
    \midrule
    CIFAR-10 & 10 & 45,000 / 5,000 / 10,000 & 10.00 & M1 / M2 / R & 82.36 / 82.28 / 83.95 \\
    KMNIST & 10 & 54,000 / 6,000 / 10,000 & 10.00 & M1 / M2 / R & 97.84 / 97.58 / 97.95 \\
    Cats vs Dogs & 2 & 18,728 / 2,341 / 2,341 & 50.15 & M1 / M2 & 87.70 / 87.95 \\
    Fast food & 10 & 2,400 / 300 / 300 & 10.00 & M1 / M2 / R & 54.67 / 52.67 / 58.00 \\
    SAT-6 & 6 & 64,800 / 8,100 / 8,100 & 37.12 & M2 high / M2 low & 99.57 / 36.01 \\
    \bottomrule
  \end{tabular}
\end{table}

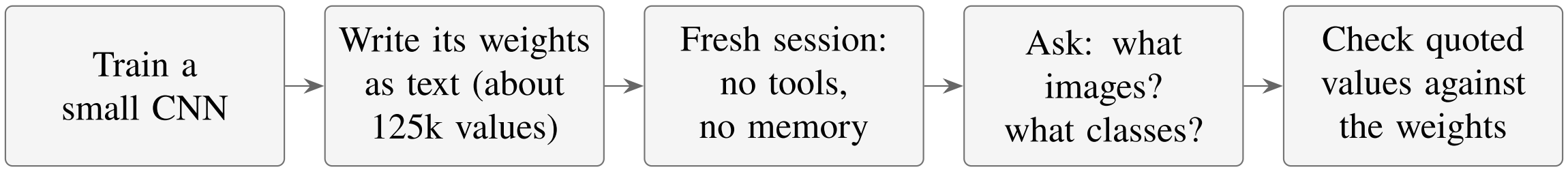
\begin{figure}[t]
  \centering
  \begin{tikzpicture}[
      font=\small,
      box/.style={draw=black!55, rounded corners=2pt, fill=black!4, text width=2.45cm, minimum height=1.55cm, align=center},
      flow/.style={-{Stealth[length=2mm]}, black!60}
    ]
    \node[box] (a) {Train a small CNN};
    \node[box, right=0.38cm of a] (b) {Write its weights as text (about 125k values)};
    \node[box, right=0.38cm of b] (c) {Fresh session: no tools, no memory};
    \node[box, right=0.38cm of c] (d) {Ask: what images? what classes?};
    \node[box, right=0.38cm of d] (e) {Check quoted values against the weights};
    \draw[flow] (a) -- (b);
    \draw[flow] (b) -- (c);
    \draw[flow] (c) -- (d);
    \draw[flow] (d) -- (e);
  \end{tikzpicture}
  \caption{Experimental pipeline. A small CNN is trained; its weights are serialized as text; a fresh session with no tools or memory receives the weights and a question; the specific values we could locate in each answer are checked against the actual parameters.}
  \label{fig:pipeline}
\end{figure}

\subsection{Prompt design}

Each prompt contains the network's layer formulas followed by every weight and batch-norm running statistic as plain text, rounded to three decimal places. The prompt omits the dataset name, class names, normalization constants, the optimizer, and augmentation. The original question asks ``Which dataset was it trained on?'', and its answer format asks for a name ``if it resembles a well-known dataset''. The revised question asks the model to describe what the images show and what each output represents, leaves the dataset name optional, and asks for separate confidence in the image type and in the classes. It also removes wording that could steer the answer toward famous datasets, colors, or the output layer. Appendix~\ref{app:prompts} reproduces both.

The wording is identical across datasets for a given architecture and output count, including the description of the input as an RGB image. This is deliberate, because changing it only for KMNIST would reveal that the dataset is grayscale. The shuffled and reordered controls also use the same wording as the unmodified prompt.

\subsection{Isolation}

Every question went to a fresh session of \texttt{claude-opus-5-5} at \texttt{xhigh} effort, through Claude Code CLI 2.1.312 in non-interactive mode, between 4 and 5 November 2026. Each session ran in a randomly named empty folder and received the prompt on standard input, with tools, MCP servers, slash commands, project files, memory, and session saving disabled. The system prompt was ``Answer the user's request.'' Claude Code also adds a short context containing the working folder, the date, the session budget, and the account email, none of which relates to the datasets.

\subsection{Verification and scoring}

We rebuilt each network directly from the prompt text and evaluated it with the prompt's written formulas, to confirm that the prompt described the intended network. Rounding changed test accuracy by at most 0.33 points. After each session, scripts checked the specific values we could locate in the response against the actual parameters, so not every claim was checked. Some claims were also tested across the whole network or on aggregate image statistics. Appendix~\ref{app:scoring} summarizes these checks.

Scoring criteria were written in the project log before sending the prompts, except for the first two \mbox{CIFAR-10} runs, which had no preset rule. The criteria were not publicly preregistered. Under the original question we recorded a pass, partial, neutral, or fail score. Under the revised question we scored image type as correct, partial, or wrong, and counted classes both in exact position and in any position, recording the dataset name separately. Appendix~\ref{app:scoring} lists the rules. Scores were assigned through Claude Code under those criteria, without an independent second rater.

\section{Results}

\subsection{CIFAR-10: recognition and label-order dependence}

\begin{figure}[t]
  \centering
  \includegraphics[width=0.85\linewidth]{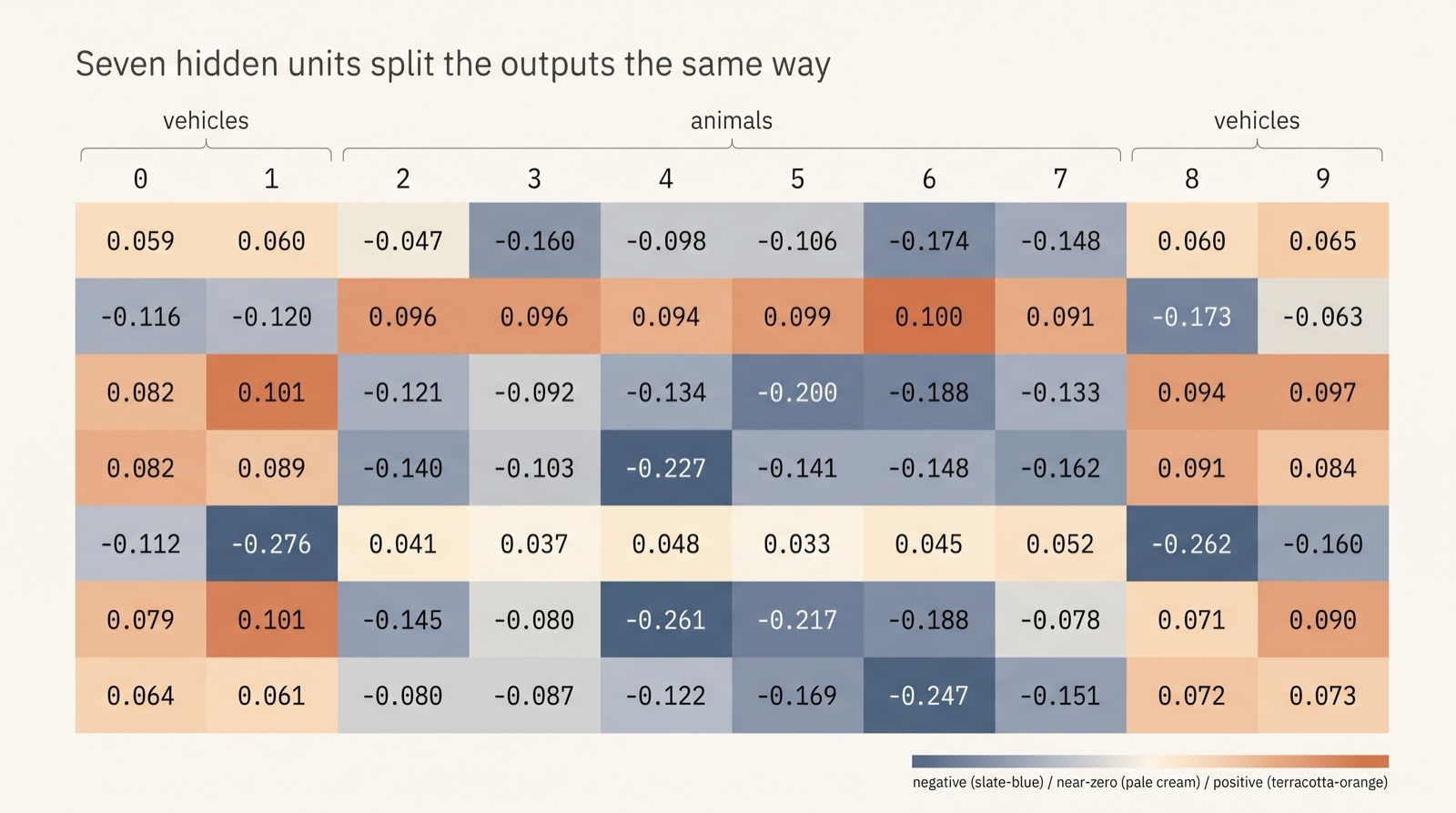}
  \caption{Output-layer structure on CIFAR-10, Model 1. Seven hidden units split outputs 0--9 into vehicles $\{0,1,8,9\}$ and animals $\{2,\dots,7\}$, matching the standard CIFAR-10 label order.}
  \label{fig:split}
\end{figure}

On CIFAR-10, both sessions named the dataset at 90\% and 92\% confidence. Their evidence is real. In both networks, the most common exact sign pattern in the output layer separates outputs $\{0,1,8,9\}$ from $\{2,3,4,5,6,7\}$, with seven hidden units each against three for the next most common pattern, and the first principal component of the output weights separates the same groups (27.7\% and 28.5\% of variance). These groups correspond to vehicles and animals in the standard label order (Figure~\ref{fig:split}).

To test whether the model derived classes from visual features or from CIFAR-10's known ordering, we ran two controls on Model 1, once each, under the original question (Table~\ref{tab:controls}). The reordered control moved the vehicle classes to outputs 2, 4, 5, and 7. The shuffled control scrambled the values within each of the 22 arrays, keeping each array's values and shape.

\begin{table}[t]
  \centering
  \caption{CIFAR-10 controls on Model 1, under the original question. Test accuracy for the two controls is measured on weights rounded as in the prompt, with labels renumbered for the reordered control. The original weights also score 82.37\% after the same rounding, against 82.36\% before it.}
  \label{tab:controls}
  \small
  \begin{tabularx}{\linewidth}{@{}lYrlr@{}}
    \toprule
    Control & Manipulation & Acc. (\%) & Model answer & Conf. (\%) \\
    \midrule
    None & Original weights & 82.36 & CIFAR-10 & 90 \\
    Reordered outputs & Vehicles moved to outputs 2, 4, 5, 7 & 82.37 & CIFAR-10 & 72 \\
    Shuffled weights & Values scrambled within each array & 7.78 & CIFAR-10 (best guess) & 45 \\
    \bottomrule
  \end{tabularx}
\end{table}

\begin{figure}[t]
  \centering
  \includegraphics[width=0.85\linewidth]{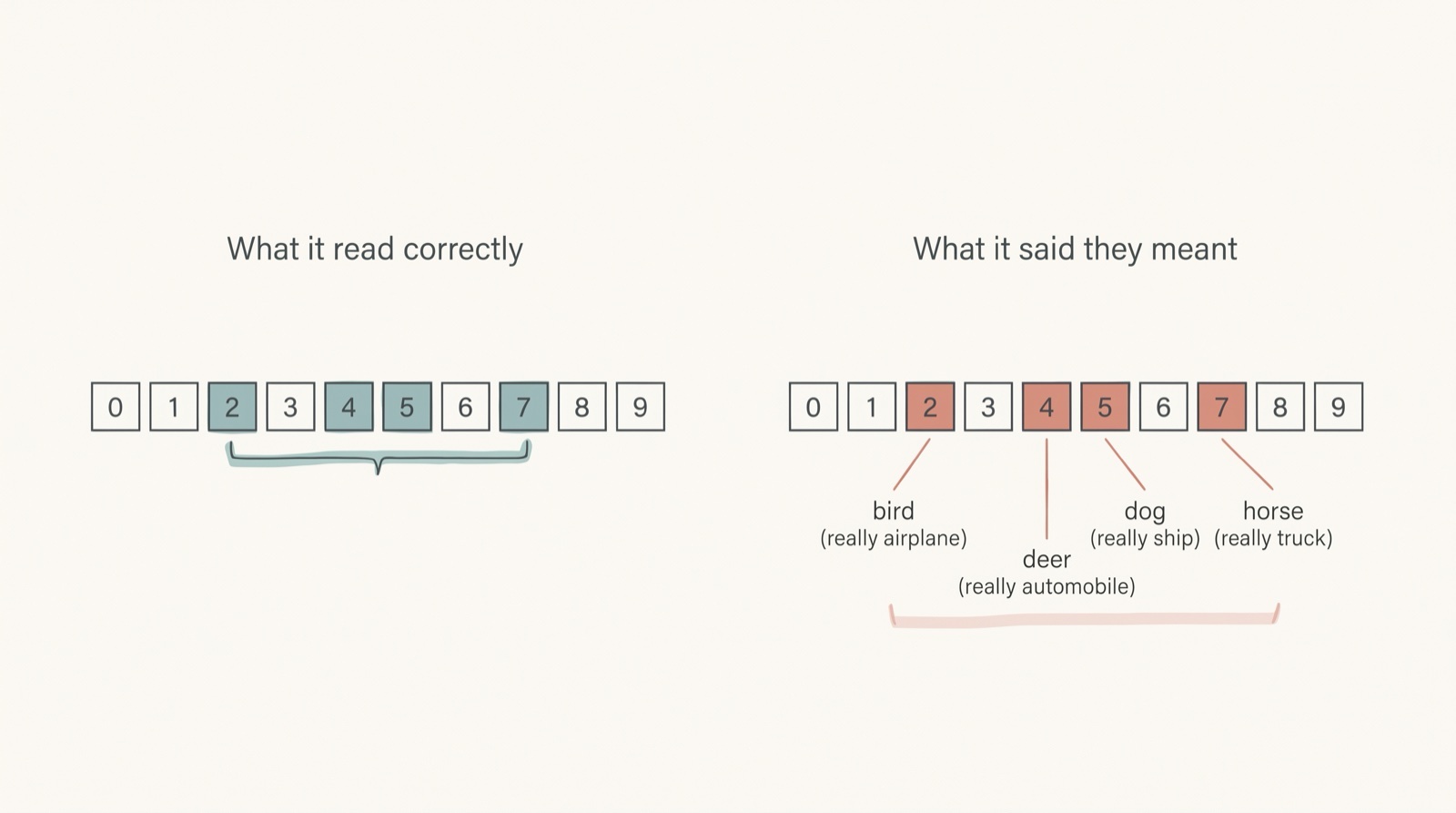}
  \caption{Reading versus naming on the reordered CIFAR-10 network. Left: the model correctly groups outputs 2, 4, 5, and 7. Right: the class names the standard order assigns to those outputs, all of which are wrong (for example, output 2 is really airplane, not bird).}
  \label{fig:naming}
\end{figure}

Permuting the output labels broke the model's class assignments. The session still found the split at the new indices (2, 4, 5, 7), which shows that it was reading the output layer's sign structure. Yet it concluded that ``the clusters match the standard label order,'' and all nine class pairs it interpreted were wrong. It read automobile and truck as deer and horse, and described outputs 4, 5, and 7, which are really automobile, ship, and truck, as quadrupeds (Figure~\ref{fig:naming}). It noticed one anomaly, an apparent anti-correlation between ``cat'' and ``dog'', and listed it as evidence against CIFAR-10, but it never considered that the labels might be reordered. A reader tracing features through the network to their visual meaning would not need the output order; a model matching clusters against a memorized CIFAR-10 template does.

On the shuffled control, confidence fell to 45\%. The session correctly reported that the vehicle/animal split was absent (a contrast of 0.0031, against 0.0103 on the trained network), and said its CIFAR-10 guess ``relies partly on it being the most common dataset.'' It did not notice that the weights had been scrambled, and it still claimed that they support ``colour natural images''. Its evidence for that, gaps between channel sums, barely separates real from scrambled weights: 3 of 16 filters show a gap above 1.0 when trained, and 2 of 16 when shuffled.

\subsection{Less-familiar datasets}
\label{sec:lessfamiliar}

Next we tested datasets outside the CIFAR-10 setup, under the original question (Table~\ref{tab:lessfamous}).

\begin{table}[t]
  \centering
  \caption{Model answers for less-familiar datasets under the original question. Confidence is in parentheses.}
  \label{tab:lessfamous}
  \small
  \begin{tabular}{@{}lll@{}}
    \toprule
    True dataset & Model 1 answer & Model 2 answer \\
    \midrule
    KMNIST (Japanese characters) & MNIST (80\%) & MNIST (85\%) \\
    Cats vs Dogs & CIFAKE (30\%) & Cats vs Dogs (35\%) \\
    Fast-food photos & CIFAR-10 (45\%) & CIFAR-10 (50\%) \\
    \bottomrule
  \end{tabular}
\end{table}

KMNIST produced the sharpest local deduction. Both sessions ruled out color datasets even though the prompt, identical across datasets, described the input as RGB. They reasoned that identical input channels receive identical gradients during training, so weight differences between channels cannot exceed the limit set by PyTorch's default initialization, 0.385. No gap exceeded it (maxima 0.367 and 0.345). The limit is not sufficient evidence of grayscale, however: Models 1 and 2 trained on fast-food color images also stay below it (maxima 0.383 and 0.382). The sessions then inferred a dark background and sparse strokes from the batch-norm statistics and named MNIST. Both estimated the background near $-0.41$ and compared it with the standard values for MNIST ($-0.42$) and KMNIST ($-0.55$). Because normalization here was computed on padded images, the true background value is $-0.47$, so this comparison was biased against KMNIST.

On Cats vs Dogs, both sessions said that the weights do not establish their answer. One inferred a red-green correlation of 0.93--0.95 (true value 0.932) and a neighboring-pixel correlation near 0.95 (true value 0.925), but then guessed CIFAKE. On the fast-food data, none of the three sessions mentioned food, including the one on the Random network, which named CIFAR-10 at 60\%. One noted that its choice ``rests on how common it is, not on the weights.'' All three noticed that the weights looked lightly trained, which fits: 95.8\% of Model 1's first-layer weights remain within their initial range.

\subsection{Architecture comparison}

We trained the Random architecture on CIFAR-10 and KMNIST and compared it with Model 1 under the revised question (Table~\ref{tab:arch}).

\begin{table}[t]
  \centering
  \caption{Architecture comparison under the revised question. ``In position'' counts outputs whose class name is correct at that index. ``Conf.'' gives confidence in the image type and in the classes.}
  \label{tab:arch}
  \small
  \setlength{\tabcolsep}{4pt}
  \begin{tabularx}{\linewidth}{@{}llrYrr@{}}
    \toprule
    Dataset & Architecture & Acc. (\%) & Image type & In position & Conf. \\
    \midrule
    CIFAR-10 & Model 1 $(16,32,64)$ & 82.36 & Correct & 10 of 10 & 95\% / 90\% \\
    CIFAR-10 & Random $(62,68,25)$ & 83.95 & Correct & 10 of 10 & 92\% / 85\% \\
    KMNIST & Model 1 $(16,32,64)$ & 97.84 & Partial: grayscale strokes, called digits & 0 of 10 & 85\% / 65\% \\
    KMNIST & Random $(62,68,25)$ & 97.95 & No answer & No answer & No answer \\
    \bottomrule
  \end{tabularx}
\end{table}

The Random network scored 10 of 10 on CIFAR-10, but this does not clear non-standard architectures. Its hidden layer widths were unconventional, but its input remained $32\times32$ and its output remained ten classes, the exact footprint of CIFAR-10. Those outer dimensions may anchor the prior: the fast-food networks share that shape and were all labeled CIFAR-10, and even a six-class SAT-6 network at $32\times32$ was read as six CIFAR-10 animal classes. Shape cannot be the whole story, though. With identical wording and the same shape, Model 1 was described as CIFAR-10 on CIFAR-10 weights and as grayscale strokes on KMNIST weights, so the weights alone changed the answer. The KMNIST Random session produced no answer: it exhausted its output limit on internal reasoning, and the automatic continuation ended in a safety stop with no stated category.

The CIFAR-10 session on Model 1 drew its own line between what it could calculate and what it was recalling: ``The weights show which outputs pair up, but not which member is which. Telling cat from dog, deer from horse, airplane from ship, or automobile from truck relies on the standard CIFAR-10 order.''

\subsection{Accuracy comparison on SAT-6}

The fast-food networks were both obscure and weakly trained, so they could not separate dataset familiarity from model quality. SAT-6 is obscure but easy to learn, which gives a more direct comparison. Table~\ref{tab:sat6} compares the full Model 2 network with its three-update counterpart, using the same prompt.

\begin{table}[t]
  \centering
  \caption{SAT-6 accuracy comparison on Model 2 under the revised question, with test accuracy and macro F1. ``Any position'' counts correct class names regardless of output index. ``Conf.'' gives confidence in the image type and in the classes.}
  \label{tab:sat6}
  \small
  \setlength{\tabcolsep}{4pt}
  \begin{tabularx}{\linewidth}{@{}lrrYrYr@{}}
    \toprule
    Network & Acc. (\%) & F1 (\%) & Image type & Any position & Name given & Conf. \\
    \midrule
    High accuracy & 99.57 & 99.51 & Wrong: animal photos & 0 of 6 & CIFAR-10, animal classes & 70\% / 45\% \\
    Low accuracy & 36.01 & 22.91 & Partial: scenes (``only a guess'') & 4 of 6 & Unknown & 30\% / 3\% \\
    \bottomrule
  \end{tabularx}
\end{table}

The near-perfect network did not lead the session to satellite imagery; it produced a more confident answer about animal photos. The low-accuracy network scored 4 of 6 classes in any position, but that came from a generic list of scene categories that the session labeled a ``guess from priors'' at 3\% confidence. This is one pair, and the low-accuracy network barely trained, with 99.1\% of its first-layer weights still in their initial range, so low accuracy and little training cannot be separated.

\begin{figure}[t]
  \centering
  \includegraphics[width=0.85\linewidth]{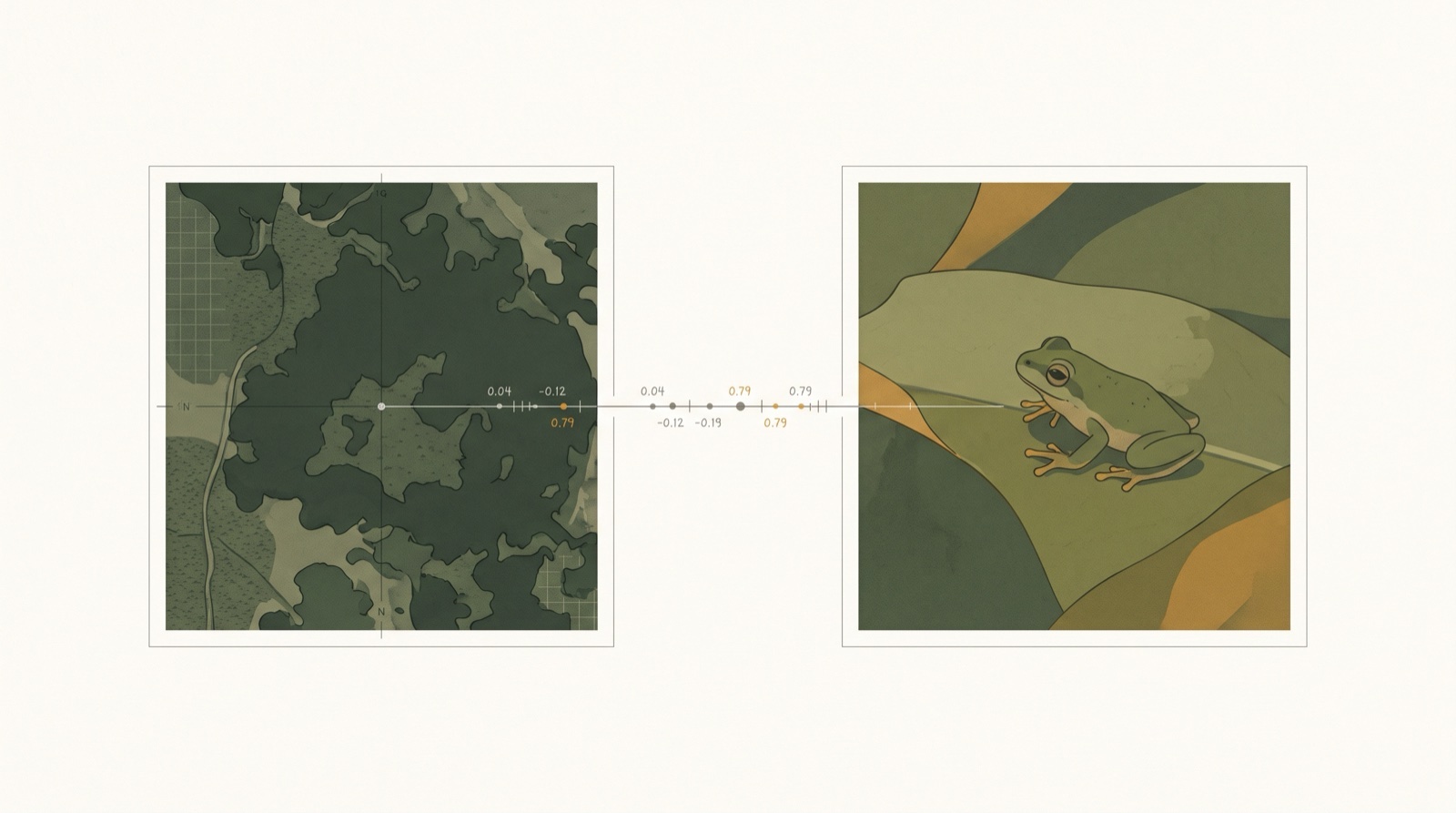}
  \caption{Same reading, wrong meaning. Conceptual illustration: a satellite forest patch and a frog are joined by a line of numbers, which are decorative and not data. The session associates a green feature with output 4, which is trees, and calls that output ``frog'', in line with its reading of the network as CIFAR-10 animals.}
  \label{fig:frog}
\end{figure}

Table~\ref{tab:sat6read} compares what the high-accuracy session extracted from the weights with the ground truth. The model found real structure, but it mapped that structure onto CIFAR-10 animals (Figure~\ref{fig:frog}).

\begin{table}[t]
  \centering
  \caption{SAT-6 high-accuracy session: what the model read, what is true in the network, and what it concluded.}
  \label{tab:sat6read}
  \small
  \begin{tabularx}{\linewidth}{@{}YYl@{}}
    \toprule
    What the session read & What is true in the network & Session's conclusion \\
    \midrule
    A ``dark green'' feature favors output 4 & It correlates with greenness (+0.64), not brightness ($-0.01$), and fires most on trees. Its top output across all hidden units is output 4, which is trees & Output 4 is \emph{frog} \\
    Outputs 1 and 3 form a strong pair (+0.54) & Outputs 1 and 3 are building and road & \emph{Cat} and \emph{dog} \\
    A ``blue'' feature favors output 0 & It correlates $-0.44$ with blueness, fires most on grassland and road, and is below average on water & Output 0 is \emph{bird} \\
    \bottomrule
  \end{tabularx}
\end{table}

\subsection{Does the model trace weights?}
\label{sec:tracing}

\begin{figure}[t]
  \centering
  \includegraphics[width=0.85\linewidth]{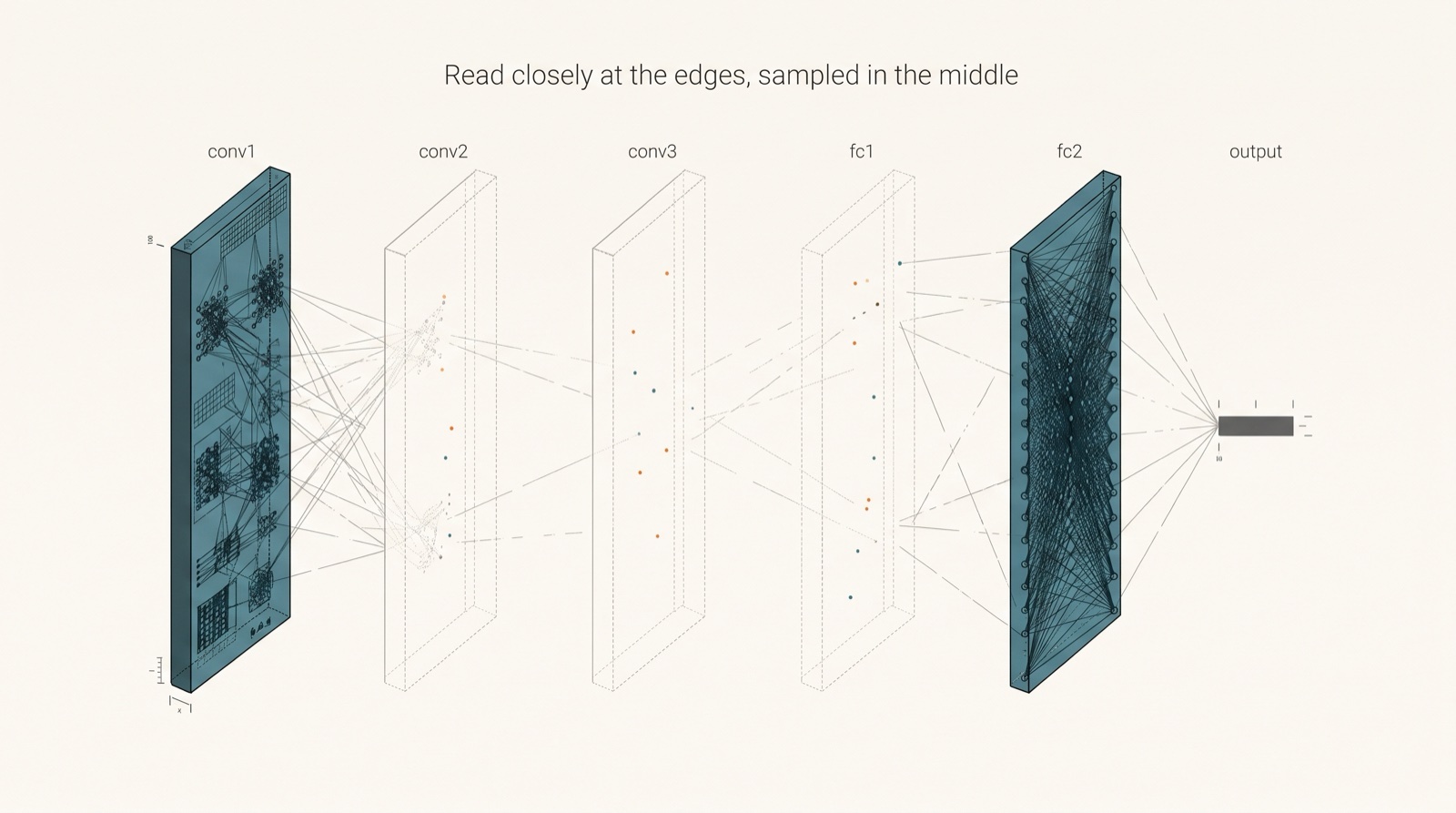}
  \caption{Where the sessions report looking. Conceptual summary, not a measurement: \texttt{conv1} and \texttt{fc2} are read closely, while \texttt{conv2}, \texttt{conv3}, and \texttt{fc1} are sampled sparsely.}
  \label{fig:coverage}
\end{figure}

Tracing a weight flow means following how signals propagate from the input pixels through every intermediate layer to the output logits. The sessions we quote here report that they did not do that:
\begin{itemize}[leftmargin=*]
  \item ``I found nothing here that identifies the dataset, and I did not trace these features through to the outputs.'' (CIFAR-10, Model 1)
  \item ``I read \texttt{conv2} only for kernel shape, did not trace \texttt{conv3} or \texttt{fc1} in detail.'' (CIFAR-10, Random)
  \item ``I worked by hand and did not trace most of conv2 and conv3.'' (KMNIST, Model 1)
  \item ``I traced class evidence for only 7 of the 77 \texttt{a3} channels.'' (Cats vs Dogs, Model 2)
  \item Rough channel-to-output effects ``using only 10 to 18 of the 78 hidden units.'' (SAT-6, Model 2)
\end{itemize}
Across these reports, the model reads the edges closely and samples the middle sparsely (Figure~\ref{fig:coverage}). These are self-reports, not measurements of an internal process, and no session was asked to compute a forward pass, so this evidence is indirect.

Claims built on partial tracing had mixed outcomes. In the SAT-6 session, three of the four channel-to-output claims named the correct top output when recomputed across all 78 hidden units, including the green feature's link to output 4. Others failed:
\begin{itemize}[leftmargin=*]
  \item \textbf{``Top-heavy'' spatial evidence.} One Cats vs Dogs session sampled 10 of the 78 hidden units, reported class evidence falling from the top row to the bottom row (about 0.33, 0.28, 0.23, 0.16), and used that to argue for ``heads and ears.'' Across all 78 hidden units, mean evidence peaks in the second row (0.109, 0.144, 0.107, 0.092).
  \item \textbf{Misread deep features.} The SAT-6 session labeled a deep feature ``blue'' and said output 0 as bird ``fits its preference for blue''. The feature's correlation with blueness is negative (Table~\ref{tab:sat6read}).
  \item \textbf{Indexing errors.} The other Cats vs Dogs session quoted real weight values from the prompt but attached them to the wrong \texttt{conv2} input channels, off by two, undermining part of its argument.
\end{itemize}

\subsection{Where the evidence lands}

Table~\ref{tab:evidence} summarizes the evidence for the components of the deflationary account. Together, these results suggest that the model can spot real statistical patterns in raw parameters, but interprets them by matching against familiar models, datasets, and framework defaults. It calculates specific local properties, such as initialization scales, optimizer step counts, and output sign patterns. When asked to synthesize those patterns into semantics, it reaches back to the templates it knows.

\begin{table}[t]
  \centering
  \caption{Evidence for the components of the deflationary account.}
  \label{tab:evidence}
  \small
  \begin{tabularx}{\linewidth}{@{}>{\raggedright\arraybackslash}p{4.2cm}>{\raggedright\arraybackslash}p{2.4cm}Y@{}}
    \toprule
    Component & Verdict & Deciding evidence \\
    \midrule
    Label order breaks it & Supported, one run & Renumbered outputs kept their remembered CIFAR-10 meanings; all nine interpreted class pairs were wrong. \\
    Input and output shape bias the guess & Supported indirectly & Same-shape fast food became CIFAR-10. Every network used $32\times32$ inputs, so input size was not varied, and KMNIST shows that shape alone does not decide the answer. \\
    It is not following weight flows & Supported indirectly & Sessions report sampling the middle layers, and several claims based on partial tracing failed our checks. No forward-pass test was run. \\
    It leans on famous datasets & Supported & KMNIST became MNIST, and the 99.57\%-accurate SAT-6 network became CIFAR-10 animal photos. \\
    Non-standard architectures break it & Unresolved & Unusual hidden layer widths did not break CIFAR-10 recognition, but input and output dimensions stayed standard. \\
    Higher accuracy helps recognition & No evidence, one pair & On SAT-6, 99.57\% accuracy came with more confidence in a wrong answer, not a correct one. \\
    Weight editing works the same way & Not tested & Editing runs were stopped before completion. \\
    \bottomrule
  \end{tabularx}
\end{table}

\section{Discussion}

\subsection{Two kinds of reading}

The results are best explained by a two-level account. At the \emph{local} level, the model performs arithmetic on the weights. Batch-norm scale parameters near one, running variances near $0.9^3=0.729$, and the sign structure of the output layer are low-dimensional statistics that can be estimated from a list of numbers. The model estimates them with few errors, and it uses framework conventions (PyTorch initialization bounds, batch-norm momentum) to turn them into claims about training. This is more than recall: finding a split at outputs 2, 4, 5, and 7 and separating KMNIST from CIFAR-10 under identical wording both require reading the numbers.

At the \emph{semantic} level, the model maps those statistics onto a small set of familiar templates. CIFAR-10 is a very common $32\times32$, ten-class benchmark, and a dark background with sparse strokes is consistent with MNIST. The same reliance on templates that yields the correct CIFAR-10 answer also yields ``frog.''

This account explains why the model can look impressive and unreliable at the same time. The local statistics are genuine and often sufficient to identify a familiar training setup. The semantic step is where the model stops reading and starts guessing. A correct dataset name is therefore weak evidence of weight-flow understanding unless the dataset's footprint is controlled.

\subsection{Why label permutation is diagnostic}

Permuting the output labels preserves the weight structure and the network's accuracy, but breaks the mapping from output indices to class names. A model that traces features from inputs to outputs could still describe the vehicle cluster as vehicles at its new indices. A model that matches a CIFAR-10 template will keep the template's order and get the names wrong. Our single reordered run shows the latter.

The shuffled control is weaker evidence because it destroys the structure that any reader would use. It mainly shows that the prior persists when the structure is absent: confidence fell, but the model still claimed to see natural images and never identified the manipulation.

\subsection{Relation to prior work}

Weight Oracles and Diff Interpretation Tuning show that models trained for the task can read weights or weight differences~\cite{goel,kabra}. Our results do not contradict those systems; they characterize a different regime. A fine-tuned oracle with an external chain of computation can be accurate on small transformers, while the general-purpose model we tested, without fine-tuning or tools, shows strong local arithmetic and prior-dominated semantics, with no evidence of flow tracing. The distinction matters when a capability is described as native to general-purpose LLMs.

Work on membership, property, and dataset inference shows that a \emph{target} model can reveal information about its training data~\cite{ganju,maini,shokri}. Our account is complementary. Our evidence suggests that the class names come largely from the \emph{interpreter's} prior over public datasets: the target network contributes the structure, and the interpreter contributes the semantics. This is also consistent with evidence that LLMs memorize training data and are partly calibrated about their knowledge~\cite{carliniExtract,carliniQuant,kadavath}.

\subsection{Confidence is not evidence}

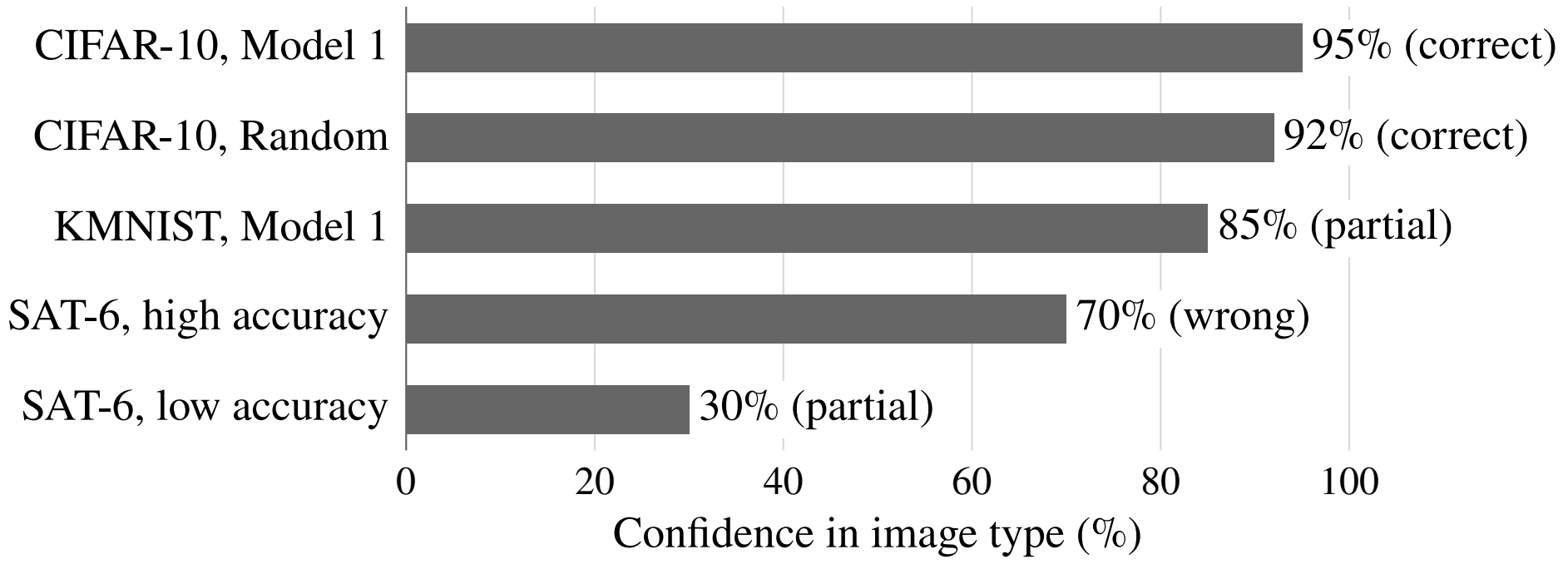
\begin{figure}[t]
  \centering
  \begin{tikzpicture}[x=0.075cm, y=0.72cm, font=\small]
    \foreach \tick in {0,20,40,60,80,100} {
      \draw[black!15] (\tick,0.45) -- (\tick,-4.45);
      \node[below, font=\footnotesize] at (\tick,-4.45) {\tick};
    }
    \foreach \row/\name/\val/\result in {
      0/{CIFAR-10, Model 1}/95/correct,
      1/{CIFAR-10, Random}/92/correct,
      2/{KMNIST, Model 1}/85/partial,
      3/{SAT-6, high accuracy}/70/wrong,
      4/{SAT-6, low accuracy}/30/partial} {
      \node[left] at (0,-\row) {\name};
      \fill[black!60] (0,-\row-0.27) rectangle (\val,-\row+0.27);
      \node[right, fill=white, inner sep=2pt] at (\val,-\row) {\val\% (\result)};
    }
    \draw[black!60] (0,0.45) -- (0,-4.45);
    \node[below] at (50,-5.0) {Confidence in image type (\%)};
  \end{tikzpicture}
  \caption{Confidence in image type under the revised question, labeled with each answer's score. The SAT-6 high-accuracy session reports 70\% for a wrong image type.}
  \label{fig:confidence}
\end{figure}

Confidence was self-reported on a 0--100 scale. It moved sensibly in several cases: 90\%, 72\%, and 45\% across the CIFAR-10 controls, 30--35\% on Cats vs Dogs, and 30\%/3\% and 10\%/1\% (image type/classes) on the three-update SAT-6 networks. It was also high on wrong answers: 80--85\% for MNIST on KMNIST networks, and 70\% for animal photos on SAT-6 (Figure~\ref{fig:confidence}). With this few runs we cannot estimate calibration. Stated confidence should therefore not be treated as evidence of understanding without controls for familiarity.

\subsection{Implications}

For weight-space programming, the reading claim should be evaluated with controls for dataset familiarity and label order. An editing result can be real even if the reading account is prior-driven, but an edit that improves a benchmark should be shown to depend on the claimed understanding rather than on search over weights. Our protocol provides a cheap first check: permute the labels, withhold the dataset name, and vary the dataset's familiarity and the network's accuracy.

For interpretability, local statistics at the edges of a network can support useful audits, such as detecting training steps, framework defaults, or output correlations, without amounting to circuit-level understanding. The distinction predicts where the model will succeed (low-dimensional statistics at the boundary) and where it will struggle (multi-layer feature tracing and obscure datasets).

For evaluation, the results argue for reporting the dataset's familiarity and the possibility of template matching. A benchmark that rewards naming a famous dataset may reward memorization rather than weight reading.

\section{Limitations}

\paragraph{Model coverage.} We test Claude Opus 5.5 only, at one effort level. The results may not transfer across models.

\paragraph{Sample size.} Most configurations were run once, on a single random seed. The counts in the tables are descriptive; we report no confidence intervals or significance tests for the model's answers, and the small sample limits the strength of comparative claims.

\paragraph{No tools, and indirect evidence on tracing.} Sessions had no calculator or code, and following about 125,000 weights by hand within one reply may be infeasible for any reader. Our results therefore show what the model did under these conditions, not that it could never trace weights. Our evidence is behavioral and relies partly on self-reports. We do not inspect the model's internal representations, and the ``prior'' language is an interpretation of the pattern, not a localized mechanism.

\paragraph{Prompt confounds.} The original question invited a ``well-known dataset'' name and the revised one does not, so answers under the two are not directly comparable. The label permutation, the shuffled control, Cats vs Dogs, and the fast-food networks ran only under the original question. The description of the input as RGB is untrue for KMNIST, and every network used $32\times32$ inputs, so the shape-prior explanation is supported only indirectly.

\paragraph{Preprocessing.} Normalization computed on padded KMNIST images shifted its background value toward MNIST's standard value, which may have contributed to the MNIST answers (Section~\ref{sec:lessfamiliar}).

\paragraph{Weak and low-accuracy networks.} The fast-food networks trained on only 2,400 images, and their test scores are noisy. The SAT-6 low-accuracy network stopped after three updates, so the accuracy comparison also compares amounts of training, and it rests on one Model 2 pair. The Model 1 result is noted in Appendix~\ref{app:results}.

\paragraph{Token limits.} Three of the sessions reported here hit the 128,000-token output limit and received an automated resume message: the shuffled CIFAR-10 run, the KMNIST Random run, and the high-accuracy SAT-6 run. Those conversations therefore differ from the single-request sessions.

\paragraph{Scoring.} Scores were assigned without an independent second rater, so we cannot report inter-rater reliability. Rule-based counts can also mislead: the SAT-6 low-accuracy network's 4 of 6 reflects a generic guess.

\paragraph{Dataset provenance.} We recorded dataset sources and file sizes but not repository commit hashes for the Hugging Face copies. The Cats vs Dogs and fast-food dataset cards state no license.

\paragraph{Model and API drift.} The model is a proprietary service. Its behavior can change across versions and sampling settings, so the exact transcripts may not reproduce.

\section{Future Work}

The following tests would separate local arithmetic from weight-flow tracing more directly than the present design.

\begin{enumerate}[leftmargin=*]
  \item \textbf{A forward-pass flow test.} Give the model a network and a single concrete input tensor, and ask for the exact output logits. A model that genuinely traces weight flows should get close; a local pattern-matcher should not. This is the most direct test of the central claim.
  \item \textbf{A shape and order test.} Train a random architecture on CIFAR-10 resized to a non-standard input resolution with permuted class labels. If the model still identifies vehicles and animals and assigns the right class names, it is not relying on shape or label order.
  \item \textbf{Non-default training hyperparameters.} Initialize weights with a non-standard distribution and change the batch-norm momentum. If the model still reconstructs the training dynamics, it is not merely leaning on PyTorch defaults.
  \item \textbf{Tool use.} Give the model Python in the same isolated setup. If it writes code to trace intermediate feature maps and correctly identifies obscure datasets such as SAT-6, the bottleneck is mental arithmetic rather than weight-space comprehension.
  \item \textbf{Blind weight editing.} Test whether the model can improve accuracy on an unseen network without gradient descent or training data. An edit that succeeds without a search over weights would be stronger evidence of understanding.
  \item \textbf{Repeats across seeds and models.} Run multiple seeds per condition, repeat the label permutation under the revised question, and include GPT-6 Astra, Claude Fable, and open-weight models, to test whether this profile is model-specific or general.
\end{enumerate}

Public preregistration of the scoring rules and prompts, a second rater, and calibration reporting would strengthen any of these tests.

\section{Conclusion}

We audited the reading side of weight-space programming with a general-purpose LLM and small convolutional networks. The model performs real local arithmetic at the edges of a network: it infers optimizer steps from batch-norm statistics and framework defaults, recovers output-layer sign patterns, and estimates weight statistics with few errors. We found no evidence that it traces the intermediate computations connecting inputs to outputs, and several of its claims from partial tracing fail verification, although a direct forward-pass test remains to be run. Its semantic judgments lean on familiar datasets, producing errors on unfamiliar data, sometimes with high confidence, and a striking failure on a label-permuted CIFAR-10 network: the arithmetic survives, but the class names do not.

The results also indicate that LLMs possess some mechanistic interpretability capability. The model recovers genuine structure from weights, including optimizer-step statistics and output correlations. The evidence suggests that this capability is supported by large-scale memory and pattern recognition over familiar training conventions, rather than by a general ability to trace the computations of an arbitrary network.

These findings do not refute Takehi et al.~\cite{takehi}; they define the conditions under which a reading claim is informative. A correct dataset name is weak evidence of weight-flow understanding when the dataset is famous and the input and output shape is its footprint. Controls for label order, shape, dataset familiarity, and accuracy help distinguish local arithmetic from prior-driven naming. More broadly, reading weights is not a single capability. It is a mixture of local statistics, framework conventions, and dataset priors, and audits of weight-space claims should report which mixture they are measuring.

\section*{AI Use Disclosure}

Claude Code (Anthropic) implemented the training and verification pipeline, helped draft the prompts and scoring criteria, assigned scores under those criteria, and assisted in preparing this manuscript. The model under test, Claude Opus 5.5, is also from Anthropic. The tested sessions had no access to that work.

\section*{Data and Code Availability}

Prompts, session transcripts, scoring logs, trained checkpoints, and code are available from the author on request.

\clearpage
\appendix

\section{Full Results}
\label{app:results}

\begin{table}[H]
  \centering
  \caption{Original question: naming the dataset. An asterisk marks the two runs that had no preset scoring rule.}
  \small
  \begin{tabular}{@{}llrlrl@{}}
    \toprule
    True dataset & Network & Acc. (\%) & Model answer & Conf. (\%) & Score \\
    \midrule
    CIFAR-10 & Model 1 & 82.36 & CIFAR-10 & 90 & Correct* \\
    CIFAR-10 & Model 2 & 82.28 & CIFAR-10 & 92 & Correct* \\
    CIFAR-10 (reordered) & Model 1 & 82.37 & CIFAR-10 & 72 & Fail \\
    CIFAR-10 (shuffled) & Model 1 & 7.78 & CIFAR-10 (best guess) & 45 & Pass \\
    KMNIST & Model 1 & 97.84 & MNIST & 80 & Partial \\
    KMNIST & Model 2 & 97.58 & MNIST & 85 & Partial \\
    Cats vs Dogs & Model 1 & 87.70 & CIFAKE & 30 & Neutral \\
    Cats vs Dogs & Model 2 & 87.95 & Cats vs Dogs & 35 & Neutral \\
    Fast food & Model 1 & 54.67 & CIFAR-10 & 45 & Neutral \\
    Fast food & Model 2 & 52.67 & CIFAR-10 & 50 & Fail \\
    Fast food & Random & 58.00 & CIFAR-10 & 60 & Fail \\
    \bottomrule
  \end{tabular}
\end{table}

\begin{table}[H]
  \centering
  \caption{Revised question: describing the dataset. ``In position'' counts outputs whose class name is correct at that index; ``any position'' counts correct names regardless of index.}
  \small
  \setlength{\tabcolsep}{4pt}
  \begin{tabular}{@{}llrlrrl@{}}
    \toprule
    True dataset & Network & Acc. (\%) & Image type & In position & Any position & Name given \\
    \midrule
    CIFAR-10 & Model 1 & 82.36 & Correct & 10 of 10 & 10 of 10 & CIFAR-10 \\
    CIFAR-10 & Random & 83.95 & Correct & 10 of 10 & 10 of 10 & CIFAR-10 \\
    KMNIST & Model 1 & 97.84 & Partial & 0 of 10 & 0 of 10 & MNIST \\
    KMNIST & Random & 97.95 & No answer & No answer & No answer & No answer \\
    SAT-6 & Model 2 high & 99.57 & Wrong & 0 of 6 & 0 of 6 & CIFAR-10, animals \\
    SAT-6 & Model 2 low & 36.01 & Partial & 0 of 6 & 4 of 6 & Unknown \\
    \bottomrule
  \end{tabular}
\end{table}

\paragraph{Note on SAT-6 Model 1.} We also ran SAT-6 Model 1 at low accuracy (63.36\%). It declined to answer, listed every class as unknown at 1\% confidence, and correctly inferred about three Adam steps. Its high-accuracy counterpart was stopped before completion, so Model 1 is omitted from the SAT-6 comparison.

\section{The Prompts}
\label{app:prompts}

Listing~\ref{lst:revised} gives the revised prompt for SAT-6 Model 2. Other networks differ only in channel counts, parameter counts, and output dimensions. Listing~\ref{lst:original} gives the question block of the original prompt, which replaces the question block of Listing~\ref{lst:revised}; the original prompt's network section and weight format are otherwise the same.

\begin{lstlisting}[caption={The revised prompt. The placeholder weight blocks are filled with the network's values when the prompt is built.},label={lst:revised}]
THE INPUT
The network sees a 32 x 32 RGB image. Rows are numbered 0 (top) to 31 (bottom) and columns 0 (left) to 31 (right). Input channel 0 is red, channel 1 is green and channel 2 is blue. Before entering the network, each channel was standardized by subtracting a fixed mean and dividing by a fixed standard deviation. Those constants are not given here.

THE OUTPUTS
The network outputs 6 numbers (logits), one per class. The class probabilities are the softmax of these logits. What each output stands for is not given.

THE NETWORK
A convolutional image classifier. Every convolution has a 3 x 3 kernel, stride 1 and padding 1. Every max pool has a 2 x 2 window and stride 2.
  a1 = bn1(maxpool(relu(conv1(x))))          conv1: 3 to 13 channels, a1 is 13 x 16 x 16
  a2 = bn2(maxpool(relu(conv2(a1))))         conv2: 13 to 33 channels, a2 is 33 x 8 x 8
  a3 = bn3(maxpool(relu(conv3(a2))))         conv3: 33 to 77 channels, a3 is 77 x 4 x 4
  f = flatten(a3)                            1,232 numbers
  h = relu(f @ fc1.weight + fc1.bias)        fc1.weight: 1,232 x 78, fc1.bias: 78
  logits = h @ fc2.weight + fc2.bias         fc2.weight: 78 x 6, fc2.bias: 6
A convolution's weight has the shape (filter, input channel, kernel row, kernel column).
Batch norm is applied per channel, using fixed statistics: bn(a) = (a - running_mean) / sqrt(running_var + 0.00001) * weight + bias.
Flattening follows f[i], where i = channel * 16 + row * 4 + column.
During training only, dropout with p = 0.3 was applied to h.
The network has 124,098 trained parameters and 246 batch norm running statistics. It was trained with supervised learning to classify images from one image dataset into 6 classes.

QUESTION. The weights below belong to one image classifier that was trained with supervised learning on one image dataset. Work from the weights to describe that dataset: what kind of images it contains and what its 6 classes are. A dataset name is optional. If the weights do not support an answer, say so and give a low confidence. You have no tools: no code execution, no calculator, nothing to run. This is a single response.

Answer in this order:
Observations: <what you see in the weights, layer by layer; quote the weights you rely on>
Reasoning: <which kinds of images and classes fit these observations, which are ruled out, and why>
Image type: <what the images show, in one or two sentences>
Classes: <one line per output, from output 0 to output 5, giving what you think it stands for>
Dataset name: <a name if you have one, otherwise unknown>
Confidence in image type: <0-100>%
Confidence in classes: <0-100>%

THE NETWORK'S WEIGHTS
Values rounded to 3 decimals.

conv1.weight, shape 13 x 3 x 3 x 3: one line per (filter, input channel), 9 values per line, in kernel row order
{{conv1.weight}}

conv1.bias, shape 13:
{{conv1.bias}}

(The same format continues for bn1, conv2, bn2, conv3, bn3, fc1 and fc2.)
\end{lstlisting}

\begin{lstlisting}[caption={The question block of the original prompt, for a 10-output network. Networks with 2 outputs use that number instead.},label={lst:original}]
QUESTION. The weights below belong to one image classifier that was trained with supervised learning on one image dataset. Which dataset was it trained on? Work from the weights. An image size of 32 x 32 with 10 outputs is common to several datasets, so base your answer on what the weights show, not on those numbers alone. If the weights do not support a specific answer, say so and give a low confidence. You have no tools: no code execution, no calculator, nothing to run. This is a single response.

Answer in this order:
Observations: <what you see in the weights: for the first convolution, what colours and patterns its filters respond to; what the later layers and batch norm statistics show; what the output layer and its biases show about the classes; quote the weights you rely on>
Reasoning: <which kinds of image dataset fit these observations, which are ruled out, and why>
Dataset: <the dataset, as specifically as you can; if it resembles a well-known dataset, give that dataset's name>
Description: <what the images show and what the 10 classes are, in one or two sentences>
Confidence: <0-100>%
\end{lstlisting}

\section{Scoring Rules and Verification}
\label{app:scoring}

\paragraph{Original question.}
\begin{itemize}[leftmargin=*]
  \item \textbf{Reordered outputs.} Pass if the session reports the split at $\{2,4,5,7\}$ and does not claim the standard label order fits. Fail if it claims a $\{0,1,8,9\}$ split or says the standard order matches.
  \item \textbf{Shuffled weights.} Pass if it says the weights do not support a specific dataset, or gives under 50\% confidence. Fail if it names a specific dataset at 50\% or more.
  \item \textbf{KMNIST.} Full pass for naming KMNIST or Kuzushiji. Partial for identifying grayscale copied to three channels, or handwritten characters or digits. Neutral for saying the weights do not support a specific dataset, or naming a color dataset under 50\%. Fail for naming a color dataset at 50\% or more.
  \item \textbf{Cats vs Dogs.} Full pass for naming Cats vs Dogs, or saying the classes are cats and dogs. Partial for saying the classes are animals or pets. Neutral for saying the weights do not support a specific dataset, naming one under 50\%, or only saying natural color photos. Fail for naming a dataset with different classes at 50\% or more. Model 2's answer met both the full-pass and the neutral rule, and we applied the stricter reading.
  \item \textbf{Fast food, first match wins.} Fail for naming a specific non-food dataset, or concluding grayscale, at 50\% or more. Full pass for saying fast food at 50\% or more. Partial for saying food images at any confidence, including fast food under 50\%. Neutral otherwise.
\end{itemize}

\paragraph{Revised question.}
\begin{itemize}[leftmargin=*]
  \item \textbf{SAT-6 image type.} Correct for satellite, aerial, overhead, remote-sensing, or land-cover imagery. Partial for outdoor scenes or landscapes without an overhead view.
  \item \textbf{CIFAR-10 image type.} Correct for photos of everyday objects, animals, and vehicles. Partial for generic natural color photos.
  \item \textbf{KMNIST image type.} Correct for handwritten characters, or Japanese, cursive, or Kuzushiji script. Partial for handwritten digits, or grayscale strokes on a dark background.
  \item \textbf{Classes.} Counted in position and in any position, with synonyms allowed. Digits score 0 on KMNIST.
  \item \textbf{Dataset name.} Recorded, not scored.
\end{itemize}

\paragraph{Verification.} Table~\ref{tab:verification} lists the sessions for which we kept tallies. An item is a quoted value or a small group of related values. Checks on the shuffled control, the original-question KMNIST runs, and the Cats vs Dogs runs were not tallied; apart from rounding-level differences and small hand-arithmetic slips, the only mismatches there were the two input indices described in Section~\ref{sec:tracing}.

\begin{table}[H]
  \centering
  \caption{Tallies of quoted items checked against the weights.}
  \label{tab:verification}
  \small
  \begin{tabularx}{\linewidth}{@{}lrrY@{}}
    \toprule
    Session & Checked & Mismatched & Note \\
    \midrule
    CIFAR-10, Model 1 (original) & 18 & 0 & One sign count was 62 against ``about 61'' \\
    CIFAR-10, Model 2 (original) & 34 & 0 & \\
    CIFAR-10, reordered & 21 & 0 & Similarity estimates within 0.03 \\
    Fast food, Model 1 & 12 & 2 & One wrong filter total ($+0.09$ against $-0.006$); correlation estimates up to 0.02 off \\
    Fast food, Model 2 & 19 & 0 & \\
    Fast food, Random & 30 & 0 & \\
    CIFAR-10, Model 1 (revised) & 6 & 0 & \\
    CIFAR-10, Random (revised) & 6 & 0 & \\
    KMNIST, Model 1 (revised) & 3 & 0 & Values correct; a written formula had a sign slip \\
    SAT-6, Model 2 high accuracy & 2 & 0 & Feature labels and channel effects assessed separately (Section~\ref{sec:tracing}) \\
    SAT-6, Model 2 low accuracy & 4 & 0 & \\
    SAT-6, Model 1 low accuracy & 9 & 0 & \\
    \bottomrule
  \end{tabularx}
\end{table}

\end{document}